\documentclass[11pt]{article}

\usepackage[letterpaper,margin=0.85in]{geometry}
\usepackage[T1]{fontenc}
\usepackage[utf8]{inputenc}
\usepackage{lmodern}
\usepackage{microtype}
\usepackage{graphicx}
\usepackage{booktabs}
\usepackage{array}
\usepackage{tabularx}
\usepackage{longtable}
\usepackage{enumitem}
\usepackage{caption}
\usepackage[hidelinks]{hyperref}
\usepackage{float}

\graphicspath{{conf_gpu_paper_figures/}}
\setlist[itemize]{leftmargin=1.4em,itemsep=0.25em,topsep=0.25em}

\title{Benchmarking Confidential GPU Inference on NVIDIA H100 under Intel TDX}
\author{%
\begin{tabular}{ccc}
Wei Wang & Abdul Hyee Waqas & Burns Smith \\
Anonym, Mozilla & Anonym, Mozilla & Anonym, Mozilla \\
\texttt{wewang@mozilla.com} & \texttt{awaqas@mozilla.com} & \texttt{busmith@mozilla.com}
\end{tabular}%
}
\date{}

\begin{document}
\maketitle

\begin{abstract}
Confidential computing is becoming a practical deployment requirement for AI inference workloads that process sensitive inputs or protect proprietary model assets. However, the performance cost of enabling confidential execution for GPU-accelerated large language model serving remains workload dependent and operationally important. This paper presents a benchmark study comparing standard non-confidential execution with confidential computing mode on a single NVIDIA H100 80GB GPU hosted in an Intel TDX confidential instance. The evaluation uses two representative language models, Mistral-7B v0.1 and Qwen3-30B-A3B, and measures time to first token, end-to-end request latency, per-request token generation throughput, global token throughput, and closed-loop request throughput under increasing concurrency. In fixed request-rate experiments, confidential mode increases average TTFT by 21.8\% for Mistral-7B and 27.8\% for Qwen3-30B-A3B, while global token throughput drops by 17.7\% and 21.1\%, respectively. In closed-loop concurrency experiments, throughput gaps remain in the 11.5--20.2\% range, but the larger model reaches its saturation knee earlier under confidential mode. The results suggest that confidential GPU inference can retain usable throughput under load, but capacity planning must account for both the steady throughput penalty and the earlier saturation behavior observed for larger models.
\end{abstract}


\section{Introduction}
Large language model serving increasingly handles sensitive user prompts, proprietary enterprise data, private retrieval context, and confidential model outputs. For such deployments, confidential computing offers an attractive execution model because it is designed to reduce trust in the cloud host while preserving the ability to use elastic infrastructure. The practical question for deployment teams is no longer whether confidential inference is conceptually useful, but how much performance capacity must be reserved when confidential mode is enabled.

This paper focuses on that operational question. It reports a controlled benchmark of confidential and non-confidential GPU inference on the same hardware class, using two model sizes and two complementary load models. The first benchmark fixes a request-rate target for each model and compares latency and token throughput. The second benchmark keeps a fixed number of in-flight requests and measures how request throughput, TTFT, and end-to-end latency evolve as concurrency increases.

The paper makes three contributions. First, it reports concrete system-level measurements for LLM inference on an NVIDIA H100 80GB GPU under Intel TDX confidential instance mode. Second, it separates fixed-rate token metrics from closed-loop concurrency behavior, which makes the saturation point easier to identify. Third, it shows that confidential mode imposes a relatively stable throughput gap, but that the larger model can reach saturation at a lower concurrency level than its non-confidential counterpart.

\section{Related Work and Positioning}
Confidential GPU inference has recently moved from architectural discussion to practical performance evaluation. Zhu et al. evaluate confidential computing on NVIDIA Hopper GPUs for LLM inference across token lengths and identify CPU-GPU data transfer as an important source of overhead in their setup \cite{zhu2024hopper}. Martinez Ibarra et al. study confidential GPU inference under traffic patterns, scheduling policies, and model swapping, reporting that non-confidential execution has lower model-swapping latency and higher throughput in their relaxed batch serving scenario \cite{ibarra2025ccgpu}. Vendor documentation also describes the hardware and cloud platform support needed for confidential GPU execution: NVIDIA describes Hopper H100 confidential computing as protecting data and applications while in use \cite{nvidia2023hcc,nvidia2023blog}, Intel TDX provides hardware-isolated trust domains for confidential VMs \cite{inteltdx}, and Google Cloud documents confidential VM support with NVIDIA H100 GPUs on A3 instances using Intel TDX \cite{gcloudgpu}.

This paper is complementary to those studies rather than a direct reproduction. Compared with Zhu et al. \cite{zhu2024hopper}, this work emphasizes service-level behavior on a cloud confidential VM configuration and reports closed-loop concurrency saturation in addition to fixed-rate token metrics. Compared with Martinez Ibarra et al. \cite{ibarra2025ccgpu}, this work does not study active model swapping; instead, it isolates steady single-model serving behavior and quantifies how confidential mode changes the saturation knee for two model sizes. The unique contribution is therefore an operational benchmark view: how much throughput headroom and latency budget should be reserved when serving Mistral-7B and Qwen3-30B-A3B on a single H100 confidential GPU instance.

\section{Experimental Setup}
\subsection{Hardware and Software Environment}
All experiments were executed in the us-central1-a zone using an a3-highgpu-1g machine type with 26 vCPUs, 234 GB of memory, one NVIDIA H100 80GB GPU, and a 250 GB balanced persistent disk. This matches the cloud configuration class documented for confidential VMs with H100 GPUs, where Google Cloud requires the a3-highgpu-1g machine type and Intel TDX for this confidential GPU path \cite{gcloudgpu}. The boot image was \texttt{ubuntu-accelerator-2404-amd64-with-nvidia-580-v20251021}. The confidential instance type was Intel TDX. Secure Boot was disabled in this setup; therefore, this paper evaluates the measured performance behavior of the benchmarked confidential instance configuration and does not make a completeness claim about the security configuration.

\begin{table}[H]
\centering
\caption{Benchmark platform configuration.}
\label{tab:platform}
\small
\begin{tabularx}{\textwidth}{@{}lX@{}}
\toprule
\textbf{Parameter} & \textbf{Value} \\
\midrule
Cloud zone & us-central1-a \\
Machine type & a3-highgpu-1g \\
CPU / memory & 26 vCPUs / 234 GB memory \\
GPU & 1 x NVIDIA H100 80GB \\
Storage & 250 GB balanced persistent disk \\
Boot image & \texttt{ubuntu-accelerator-2404-amd64-with-nvidia-580-v20251021} \\
Confidential instance type & Intel TDX \\
Secure Boot & Disabled \\
Benchmark tool & \texttt{cc\_benchmarks} \\
\bottomrule
\end{tabularx}
\end{table}

\subsection{Models and Execution Modes}
The benchmark uses Mistral-7B v0.1 and Qwen3-30B-A3B. Mistral-7B is a 7-billion-parameter model designed for efficient inference using mechanisms such as grouped-query attention and sliding-window attention \cite{mistral7b}. Qwen3 is a model family that includes dense and mixture-of-experts architectures, including the Qwen3-30B-A3B variant used in this benchmark \cite{qwen3}. These two models provide a useful contrast between a smaller model and a larger model under the same single-GPU serving setup. Each experiment was run in two modes: standard non-confidential execution, denoted Non-CC, and confidential computing execution, denoted CC.

\subsection{Metrics}
The fixed request-rate benchmark reports time to first token (TTFT), end-to-end request latency, per-request token throughput, and global token throughput. The fixed-rate table includes both average and p50 values. Unless explicitly marked as p50, the fixed-rate discussion and Figure~\ref{fig:fixed-overheads} use average values. The closed-loop concurrency benchmark reports request throughput in requests per second, p50 TTFT, and p50 end-to-end latency. Closed-loop throughput values are trial-averaged request-throughput measurements, while closed-loop latency and TTFT values are trial-averaged p50 measurements. The uploaded benchmark output labels the fixed request-rate latency metrics in seconds and the closed-loop latency metrics in milliseconds; this paper preserves those reported units rather than normalizing or relabeling them.

For latency-style metrics, positive overhead is interpreted as the relative increase of CC over Non-CC, computed as $(\mathrm{CC}-\mathrm{Non\mbox{-}CC})/\mathrm{Non\mbox{-}CC}$. For throughput-style metrics, the paper reports the relative throughput gap, where larger values indicate that Non-CC achieves higher throughput than CC, computed as $(\mathrm{Non\mbox{-}CC}-\mathrm{CC})/\mathrm{CC}$. When CC reports a lower latency or TTFT than Non-CC at a specific concurrency point, this paper describes the result as a relative decrease rather than as negative overhead.

\subsection{Load-Generation Methodology}
Two load patterns were evaluated. In the fixed request-rate experiment, Mistral-7B was tested at 10.0 requests per second and Qwen3-30B-A3B was tested at 5.0 requests per second. In the closed-loop concurrency experiment, a new request was immediately initiated after a previous request completed, maintaining $N$ in-flight requests. For each concurrency level, throughput, p50 TTFT, and p50 latency were averaged over three trials. This closed-loop method is common in service capacity testing because it exposes saturation behavior rather than only measuring a single low-load operating point. Modern LLM serving systems are highly sensitive to request scheduling, KV-cache management, and batching behavior, which is why throughput should be interpreted together with latency and concurrency \cite{vllm}.

\section{Fixed Request-Rate Results}
The fixed request-rate benchmark shows a consistent confidential-mode latency penalty for both models. The penalty is larger for Qwen3-30B-A3B than for Mistral-7B, while the global throughput gap also increases from 17.7\% to 21.1\% as the workload moves to the larger model. The current result artifact contains average and p50 values, but it does not contain p90 or p99 tail-latency values; those should be reported in a future revision if raw request-level traces or percentile summaries are available.

\begin{table}[H]
\centering
\caption{Fixed request-rate benchmark summary.}
\label{tab:fixed-rate}
\scriptsize
\resizebox{\textwidth}{!}{%
\begin{tabular}{@{}llrrrrrrr@{}}
\toprule
\textbf{Model} & \textbf{Rate} & \textbf{Metric} & \textbf{Non-CC avg} & \textbf{CC avg} & \textbf{Avg overhead/gap} & \textbf{Non-CC p50} & \textbf{CC p50} & \textbf{p50 overhead/gap} \\
\midrule
Mistral-7B v0.1 & 10.0 req/s & TTFT & 279.56 & 340.64 & 21.8\% & 281.22 & 339.95 & 20.9\% \\
Mistral-7B v0.1 & 10.0 req/s & Request latency & 314.72 & 381.04 & 21.1\% & 316.70 & 380.97 & 20.3\% \\
Mistral-7B v0.1 & 10.0 req/s & Throughput (tok/s) & 28.50 & 25.18 & 13.2\% & 28.16 & 24.46 & 15.1\% \\
Mistral-7B v0.1 & 10.0 req/s & Global throughput (tok/s) & 1458.65 & 1239.26 & 17.7\% & N/A & N/A & N/A \\
Qwen3-30B-A3B & 5.0 req/s & TTFT & 394.61 & 504.35 & 27.8\% & 394.75 & 511.85 & 29.7\% \\
Qwen3-30B-A3B & 5.0 req/s & Request latency & 414.51 & 527.18 & 27.2\% & 414.70 & 535.28 & 29.1\% \\
Qwen3-30B-A3B & 5.0 req/s & Throughput (tok/s) & 50.20 & 43.85 & 14.5\% & 50.20 & 43.25 & 16.1\% \\
Qwen3-30B-A3B & 5.0 req/s & Global throughput (tok/s) & 993.80 & 820.31 & 21.1\% & N/A & N/A & N/A \\
\bottomrule
\end{tabular}%
}
\end{table}

\begin{figure}[H]
\centering
\includegraphics[width=0.98\textwidth]{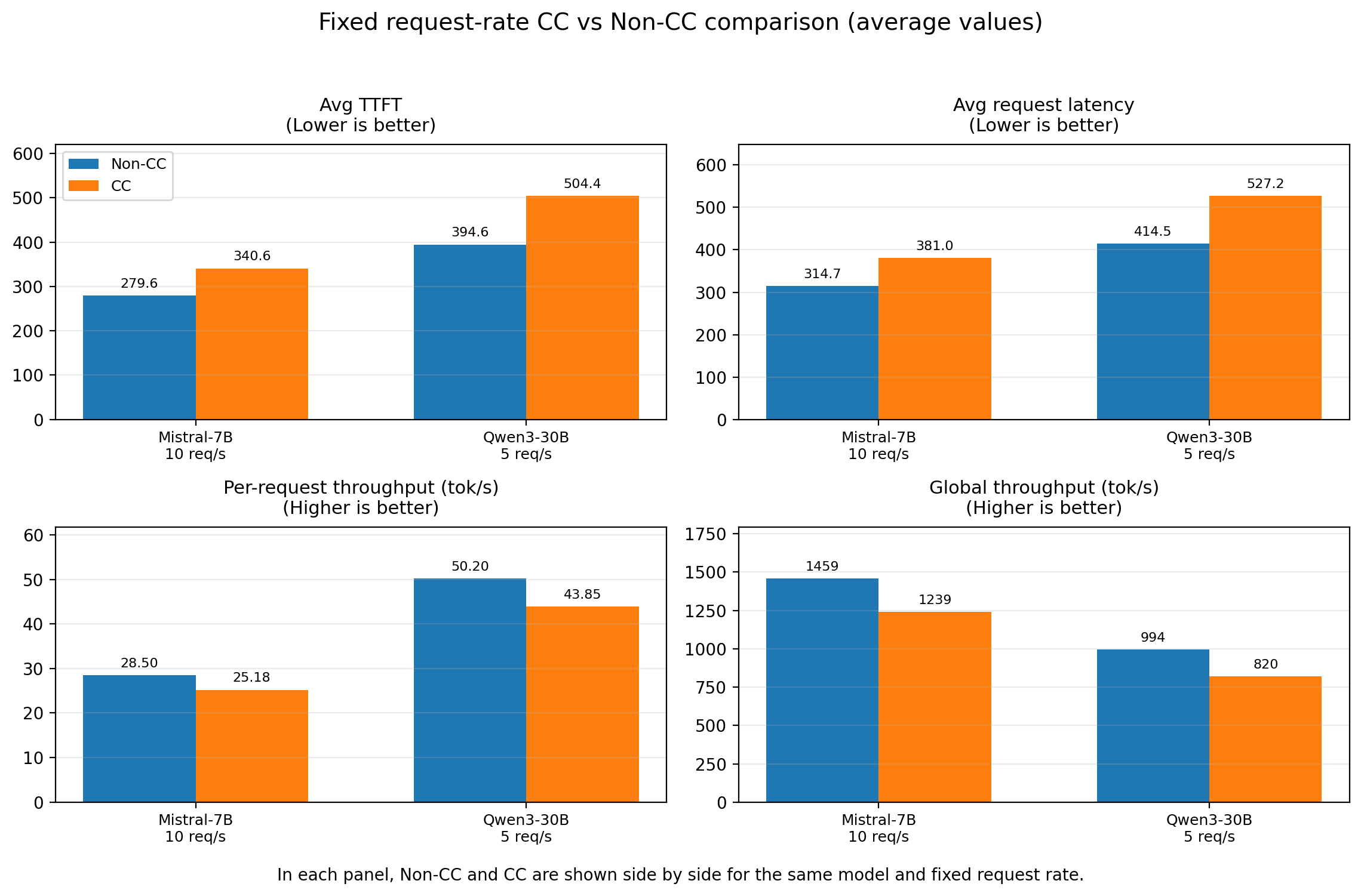}
\caption{Fixed request-rate CC versus Non-CC comparison using average values from Table~\ref{tab:fixed-rate}. The figure shows the measured CC and Non-CC values directly, not only percentage overheads. In each panel, bars are grouped by model; within each group, the left bar is Non-CC and the right bar is CC. For TTFT and request latency, lower is better. For per-request and global throughput, higher is better.}
\label{fig:fixed-overheads}
\end{figure}

To avoid ambiguity between the two execution modes, Figure~\ref{fig:fixed-overheads} shows CC and Non-CC side by side in the same fixed-rate comparison. Each panel uses average values from Table~\ref{tab:fixed-rate}. In every panel, the left bar is Non-CC and the right bar is CC for the same model and fixed request rate.

\subsection{Latency Behavior}
For Mistral-7B, confidential mode increases average TTFT from 279.56 to 340.64 and average request latency from 314.72 to 381.04. The corresponding overheads are 21.8\% and 21.1\%. For Qwen3-30B-A3B, average TTFT increases from 394.61 to 504.35, and average request latency increases from 414.51 to 527.18. The corresponding overheads are 27.8\% and 27.2\%. The larger model therefore shows a higher latency sensitivity to CC mode in this setup.

\subsection{Throughput Behavior}
The token generation throughput degradation is smaller than the TTFT degradation but remains operationally meaningful. Mistral-7B average per-request throughput drops from 28.50 tok/s to 25.18 tok/s, while global throughput drops from 1458.65 tok/s to 1239.26 tok/s. Qwen3-30B-A3B average per-request throughput drops from 50.20 tok/s to 43.85 tok/s, while global throughput drops from 993.80 tok/s to 820.31 tok/s. The fixed-rate results therefore suggest that confidential mode affects both user-visible latency and aggregate serving capacity.

\section{Closed-Loop Concurrency Results}
The closed-loop benchmark examines how each mode behaves as the number of in-flight requests increases. This is useful for identifying the saturation knee rather than only measuring performance at a fixed request rate.

\begin{figure}[H]
\centering
\includegraphics[width=0.86\textwidth]{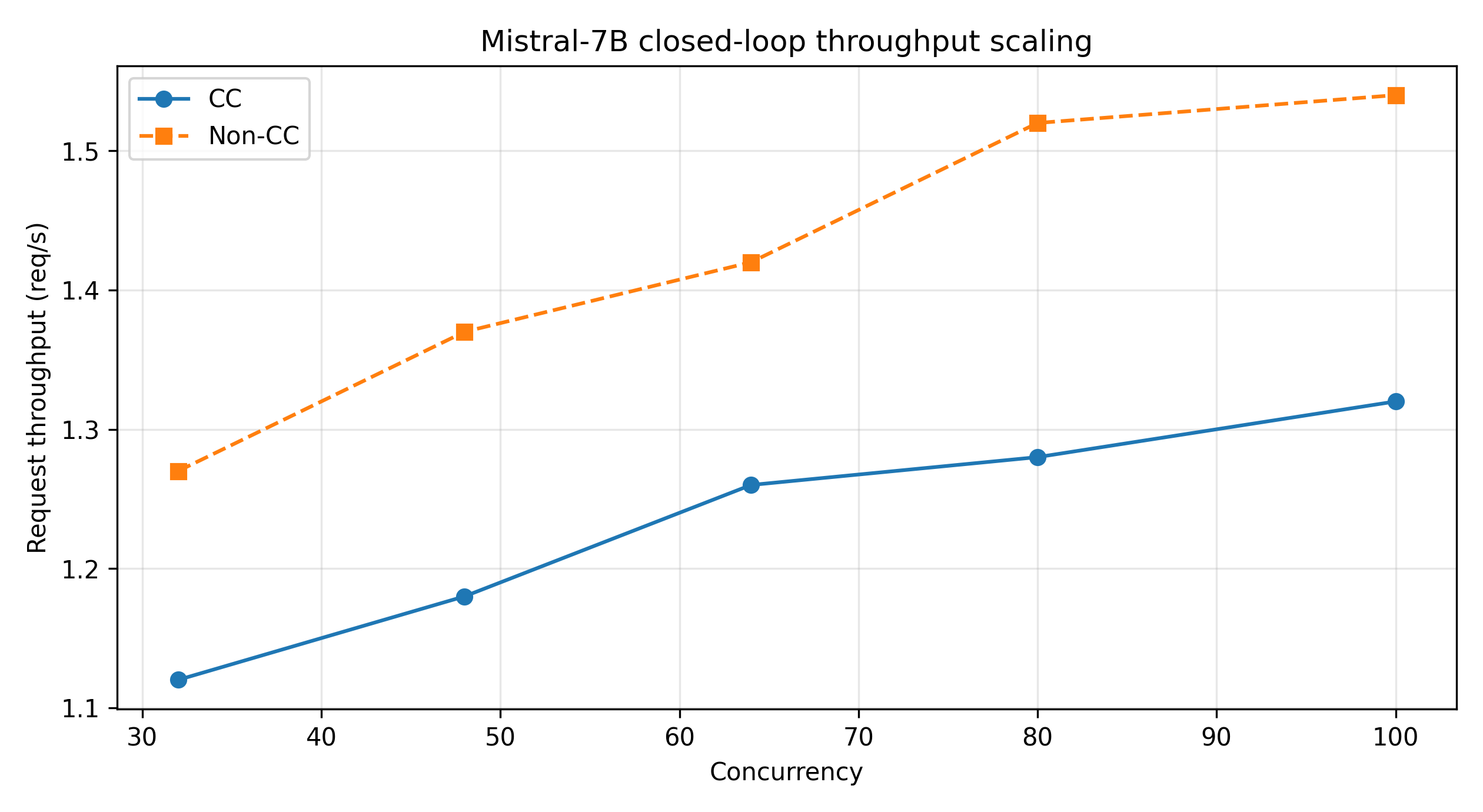}
\caption{Mistral-7B request throughput under closed-loop load. Points show trial-averaged request throughput for CC and Non-CC.}
\label{fig:mistral-throughput}
\end{figure}

\begin{figure}[H]
\centering
\includegraphics[width=0.86\textwidth]{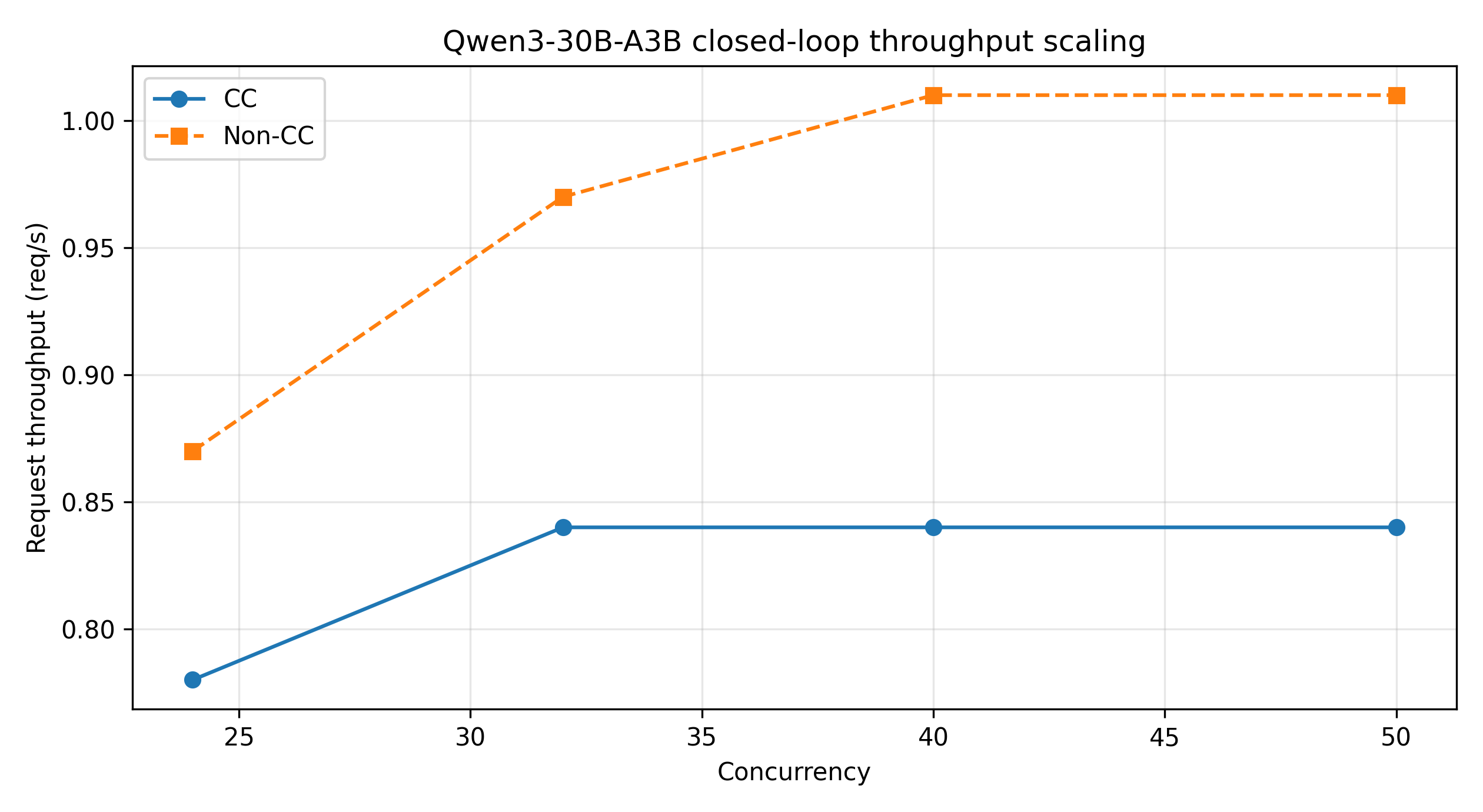}
\caption{Qwen3-30B-A3B request throughput under closed-loop load. Points show trial-averaged request throughput for CC and Non-CC.}
\label{fig:qwen-throughput}
\end{figure}

\subsection{Throughput Scaling}
Mistral-7B throughput grows with concurrency for both modes and begins to plateau in the 80--100 concurrency range. At $c=100$, CC reaches 1.32 req/s while Non-CC reaches 1.54 req/s, corresponding to a 16.7\% throughput gap. Across the tested Mistral range, the throughput gap remains between 12.7\% and 18.8\%.

Qwen3-30B-A3B shows a different shape. CC reaches 0.84 req/s by $c=32$ and remains flat at $c=40$ and $c=50$. Non-CC reaches 0.97 req/s at $c=32$ and 1.01 req/s at $c=40$ and $c=50$. This indicates that the larger model saturates earlier under CC mode. The CC throughput ceiling is lower, and the saturation knee appears at a lower concurrency level.

\subsection{TTFT Regimes}
TTFT is not monotonic across all regimes. For Mistral-7B, CC reports lower p50 TTFT than Non-CC at lower concurrency levels $c=32$ and $c=48$: CC TTFT is 19.4\% lower at $c=32$ and 23.2\% lower at $c=48$. Once the workload reaches higher concurrency, CC TTFT becomes slightly higher than Non-CC, with positive overhead from 1.7\% to 6.0\%. This suggests that TTFT should not be interpreted in isolation; it must be interpreted relative to whether the system is pre-saturation, near the knee, or post-saturation.

Qwen3-30B-A3B has one prominent saturation anomaly at $c=32$: CC p50 TTFT is 20,823.1 ms while Non-CC p50 TTFT is 2,874.1 ms. Because the two modes are at different points relative to their saturation knees at $c=32$, the TTFT comparison at that single point is not representative of steady-state overhead. At $c=40$ and $c=50$, where both modes are closer to plateau, Qwen3-30B-A3B TTFT overhead is 9.7\%.

\subsection{End-to-End Latency}
\begin{figure}[H]
\centering
\includegraphics[width=0.86\textwidth]{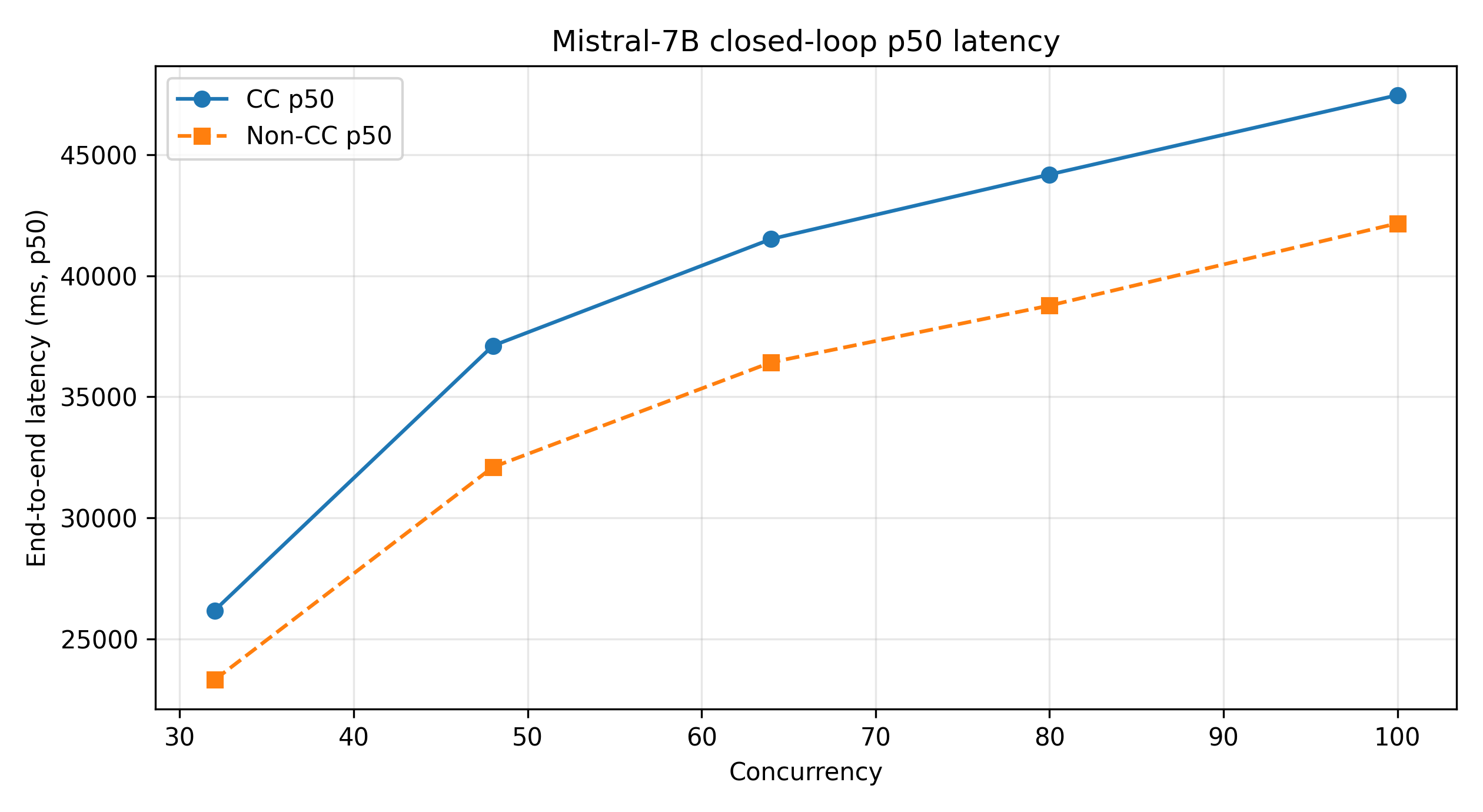}
\caption{Mistral-7B p50 end-to-end latency under closed-loop load. Points show trial-averaged p50 latency for CC and Non-CC.}
\label{fig:mistral-latency}
\end{figure}

\begin{figure}[H]
\centering
\includegraphics[width=0.86\textwidth]{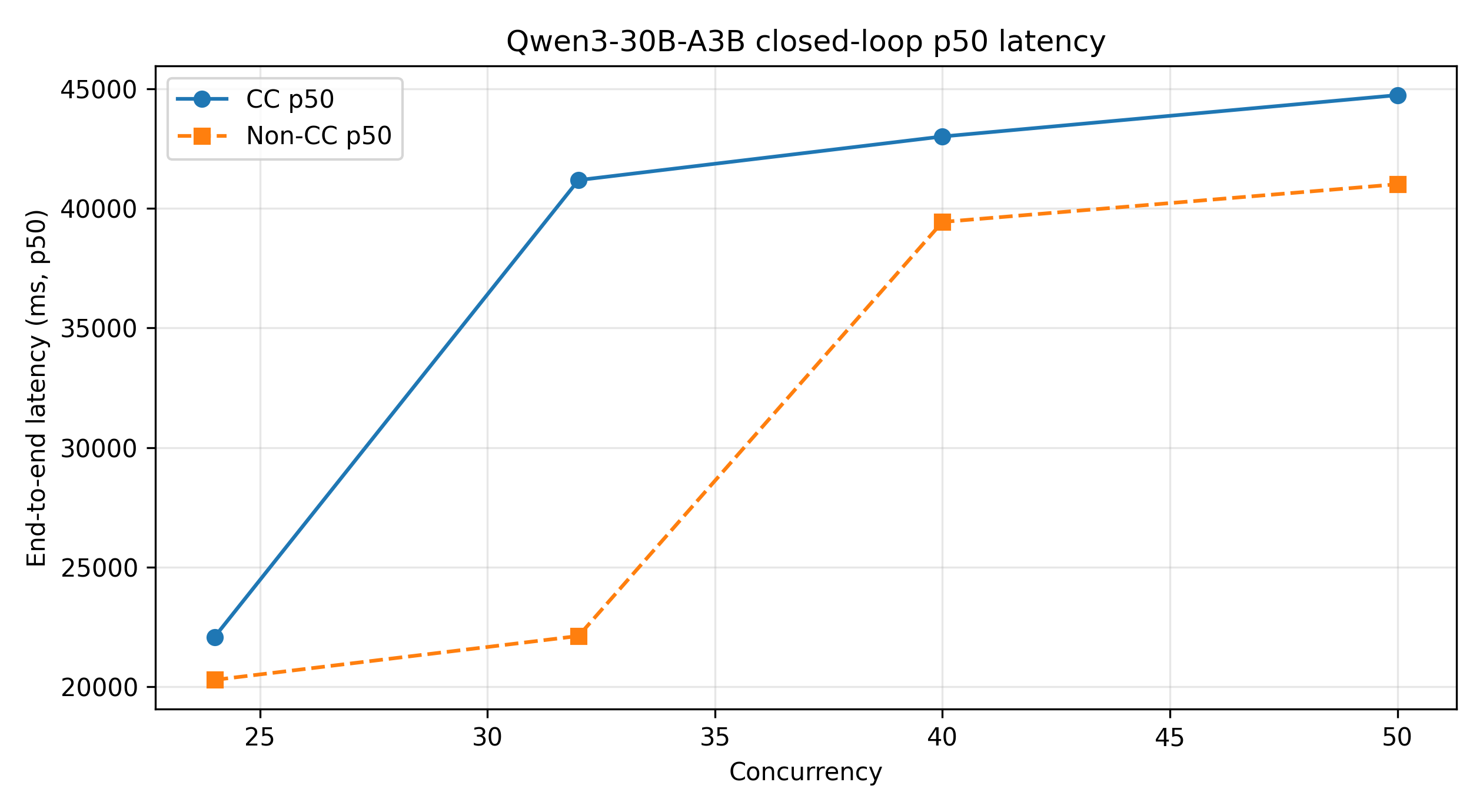}
\caption{Qwen3-30B-A3B p50 end-to-end latency under closed-loop load. Points show trial-averaged p50 latency for CC and Non-CC.}
\label{fig:qwen-latency}
\end{figure}

Mistral-7B p50 request latency is consistently higher in CC mode across all tested concurrencies. The overhead ranges from 12.3\% to 15.6\%, with no single anomaly dominating the trend. Qwen3-30B-A3B p50 latency overhead is 8.9\% at $c=24$ and approximately 9.1\% at $c=40$ and $c=50$, excluding the $c=32$ saturation anomaly. The latency curves show that confidential mode does not merely shift all values upward; it can also alter where the system transitions from efficient scaling into saturation.

\section{Discussion}
\subsection{Model-Size Sensitivity}
The most visible fixed-rate result is the higher latency overhead for Qwen3-30B-A3B relative to Mistral-7B. In this benchmark, the larger model shows TTFT and request-latency overheads near 27--30\%, while the smaller model stays near 20--22\%. This does not prove that model size alone determines overhead, because sequence length, batching behavior, runtime configuration, memory pressure, and scheduler behavior can also contribute. Nevertheless, the result is operationally meaningful: larger models may require more conservative capacity planning when confidential mode is enabled.

\subsection{Capacity Planning and Saturation Knees}
The closed-loop experiments show that confidential mode should be evaluated not only by percent overhead at a single operating point but also by the concurrency at which the service saturates. For Mistral-7B, both modes continue to scale into the 80--100 range. For Qwen3-30B-A3B, CC saturates around $c=32$ while Non-CC continues improving until approximately $c=40$. This means that a deployment sized only from low-load measurements may under-provision the confidential configuration once production traffic approaches the saturation knee.

\subsection{TTFT as an Operating-Regime Metric}
TTFT is a user-facing metric, but it can be misleading when two configurations are being compared at different points of their throughput curve. The Mistral-7B low-concurrency results show lower CC p50 TTFT than Non-CC at two points even though CC throughput remains lower. The Qwen3-30B-A3B $c=32$ result shows the opposite extreme: a very large TTFT gap caused by one mode entering saturation earlier than the other. The practical conclusion is that TTFT should be paired with throughput and queueing-state indicators before drawing conclusions about confidential-mode overhead.

\subsection{Practical Implications}
For teams deploying confidential LLM inference, the benchmark suggests four practical rules. First, reserve 15--25\% additional throughput capacity as an initial planning heuristic for similar single-H100 deployments, then validate with workload-specific traces. Second, measure the saturation knee for each target model, because the larger model can lose concurrency headroom before the smaller model. Third, separate TTFT analysis into pre-saturation and post-saturation regions. Fourth, treat confidential mode as a performance profile that must be tuned, not as a simple binary overhead multiplier.

\section{Limitations and Threats to Validity}
\begin{itemize}
    \item The benchmark covers two models and one GPU configuration. The results should not be generalized to all model sizes, inference engines, batch schedulers, or GPU counts without additional measurements.
    \item Secure Boot was disabled in the reported setup. The paper therefore focuses on performance behavior under the tested confidential instance configuration rather than claiming a fully validated production security posture.
    \item The benchmark does not include low-level profiling of PCIe transfer time, GPU utilization, memory bandwidth, kernel-level timing, or runtime scheduler internals. Those measurements would be needed to attribute overhead to a specific subsystem.
    \item The fixed request-rate benchmark and closed-loop benchmark use different reporting units and load models. The results are complementary but should not be directly merged into a single normalized curve.
    \item Only aggregate and p50 statistics are reported in the current artifact. Tail latency, especially p90, p95, and p99, request-length distribution, prompt/output token distributions, and warmup effects may materially affect production behavior.
\end{itemize}

\section{Further Research Directions}
The current results are sufficient to show a measurable confidential-mode performance gap, but additional measurements would make the root-cause analysis stronger.

\begin{itemize}
    \item \textbf{Tail-latency reporting.} If raw request traces or percentile summaries are available, future revisions should report p90 and preferably p95/p99 for TTFT and end-to-end latency. This would show whether confidential mode primarily shifts the median or also widens the tail.
    \item \textbf{Low-level profiling.} Future experiments should capture PCIe transfer time, CPU-GPU copy overhead, GPU utilization, HBM bandwidth, kernel execution time, batching behavior, and scheduler queueing time. Prior work identifies CPU-GPU transfer as a possible source of confidential-mode overhead, so these counters would help separate transfer overhead from compute and scheduling effects \cite{zhu2024hopper,nvidia2023hcc}.
    \item \textbf{Hardware sensitivity.} The benchmark should be repeated on different hardware configurations, including single-GPU versus multi-GPU setups, PCIe versus NVLink-connected systems, different CPU and memory configurations, and different confidential-VM settings. This is especially important because multi-GPU confidential deployments may introduce additional communication and peer-memory behavior.
    \item \textbf{Model and workload coverage.} Future runs should include more dense and mixture-of-experts models, additional model sizes, longer context windows, quantized variants, embedding workloads, and production-like prompt/output length distributions.
    \item \textbf{Traffic realism.} Beyond fixed-rate and closed-loop tests, future benchmarks should include bursty arrivals, mixed request lengths, warm and cold model states, and scheduler policies used in production serving stacks.
\end{itemize}

\section{Conclusion}
This paper presented a benchmark study of confidential GPU inference on an NVIDIA H100 80GB instance using Intel TDX. Across Mistral-7B and Qwen3-30B-A3B, confidential mode introduces a clear user-visible latency penalty and a stable throughput gap. In fixed request-rate experiments, TTFT and request latency overheads range from approximately 21\% for Mistral-7B to approximately 27--30\% for Qwen3-30B-A3B. Global token throughput drops by 17.7\% and 21.1\%, respectively. In closed-loop concurrency tests, throughput gaps stay within 11.5--20.2\%, but Qwen3-30B-A3B reaches saturation at a lower concurrency under CC mode. These results show that confidential GPU inference is operationally feasible on a single H100, but must be capacity-planned using model-specific saturation measurements rather than a single fixed overhead assumption. The next revision should add p90 or higher-percentile tail-latency results if the underlying benchmark data is available.

\appendix
\section{Full Closed-Loop Results}

\begin{table}[H]
\centering
\caption{Mistral-7B closed-loop concurrency results. The table is split into throughput and latency panels to fit portrait layout without reducing readability.}
\label{tab:mistral-closed-loop}
\small
\textbf{Panel A: Request throughput}\\[0.3em]
\begin{tabular}{@{}rrrr@{}}
\toprule
\textbf{Concurrency} & \textbf{CC req/s} & \textbf{Non-CC req/s} & \textbf{Gap} \\
\midrule
32  & 1.12 & 1.27 & 13.4\% \\
48  & 1.18 & 1.37 & 16.1\% \\
64  & 1.26 & 1.42 & 12.7\% \\
80  & 1.28 & 1.52 & 18.8\% \\
100 & 1.32 & 1.54 & 16.7\% \\
\bottomrule
\end{tabular}

\vspace{0.9em}
\textbf{Panel B: Median TTFT and end-to-end latency}\\[0.3em]
\scriptsize
\begin{tabular}{@{}rrrrrrr@{}}
\toprule
\textbf{Conc.} & \shortstack{\textbf{CC TTFT}\\\textbf{ms p50}} & \shortstack{\textbf{Non-CC TTFT}\\\textbf{ms p50}} & \shortstack{\textbf{CC TTFT}\\\textbf{vs Non-CC}} & \shortstack{\textbf{CC latency}\\\textbf{ms p50}} & \shortstack{\textbf{Non-CC latency}\\\textbf{ms p50}} & \shortstack{\textbf{Latency}\\\textbf{overhead}} \\
\midrule
32  & 1186.7 & 1472.0 & 19.4\% lower & 26170.9 & 23304.2 & 12.3\% \\
48  & 1218.9 & 1586.4 & 23.2\% lower & 37108.4 & 32093.7 & 15.6\% \\
64  & 3284.8 & 3217.6 & 2.1\% higher & 41516.8 & 36422.8 & 14.0\% \\
80  & 6359.8 & 6000.7 & 6.0\% higher & 44184.3 & 38768.7 & 14.0\% \\
100 & 8262.2 & 8127.9 & 1.7\% higher & 47462.1 & 42164.9 & 12.6\% \\
\bottomrule
\end{tabular}
\end{table}

\begin{table}[H]
\centering
\caption{Qwen3-30B-A3B closed-loop concurrency results. The split-panel layout keeps the saturation anomaly visible while preserving page fit.}
\label{tab:qwen-closed-loop}
\small
\textbf{Panel A: Request throughput}\\[0.3em]
\begin{tabular}{@{}rrrr@{}}
\toprule
\textbf{Concurrency} & \textbf{CC req/s} & \textbf{Non-CC req/s} & \textbf{Gap} \\
\midrule
24 & 0.78 & 0.87 & 11.5\% \\
32 & 0.84 & 0.97 & 15.5\% \\
40 & 0.84 & 1.01 & 20.2\% \\
50 & 0.84 & 1.01 & 20.2\% \\
\bottomrule
\end{tabular}

\vspace{0.9em}
\textbf{Panel B: Median TTFT and end-to-end latency}\\[0.3em]
\scriptsize
\begin{tabular}{@{}rrrrrrr@{}}
\toprule
\textbf{Conc.} & \shortstack{\textbf{CC TTFT}\\\textbf{ms p50}} & \shortstack{\textbf{Non-CC TTFT}\\\textbf{ms p50}} & \shortstack{\textbf{CC TTFT}\\\textbf{vs Non-CC}} & \shortstack{\textbf{CC latency}\\\textbf{ms p50}} & \shortstack{\textbf{Non-CC latency}\\\textbf{ms p50}} & \shortstack{\textbf{Latency}\\\textbf{overhead}} \\
\midrule
24 & 1285.0  & 1199.4  & 7.6\% higher & 22073.1 & 20277.8 & 8.9\% \\
32 & 20823.1 & 2874.1  & 725.0\% higher & 41187.0 & 22114.2 & 86.2\% \\
40 & 22180.3 & 20221.1 & 9.7\% higher & 43011.9 & 39436.7 & 9.1\% \\
50 & 23937.9 & 21803.8 & 9.7\% higher & 44742.0 & 41013.9 & 9.1\% \\
\bottomrule
\end{tabular}
\end{table}

\section{Reproducibility Notes}
The benchmark artifact records two result categories: standard mode versus confidential computing mode for fixed request-rate experiments, and QPS-impact experiments for closed-loop concurrency. The fixed request-rate experiments use request rates of 10.0 req/s for Mistral-7B and 5.0 req/s for Qwen3-30B-A3B. The closed-loop concurrency experiments average request throughput over three trials per concurrency level and report trial-averaged p50 TTFT and p50 end-to-end latency. The current artifact does not include p90 or p99 latency values.

\end{document}